\documentclass[preprint,review,12pt]{elsarticle}

\usepackage{amssymb}
\usepackage{amsmath,amsfonts}
\usepackage{algorithmic}
\usepackage{algorithm}
\usepackage{array}
\usepackage{textcomp}
\usepackage{stfloats}
\usepackage{url}
\usepackage{verbatim}
\usepackage{graphicx}
\usepackage{caption}
\usepackage{subcaption}
\usepackage{float}
\usepackage{multirow}
\usepackage{color}
\usepackage{wrapfig}

\usepackage{graphicx}
\usepackage{amsmath}
\usepackage{amssymb}
\usepackage{booktabs}
\usepackage{color}  

\usepackage{subcaption}
\usepackage{booktabs}
\usepackage{bm}
\usepackage{multirow}
\usepackage{bbding}
\usepackage{color}

\usepackage{amsthm}

\usepackage{graphicx}
\usepackage{amsmath}
\usepackage{amssymb}
\usepackage{booktabs}
\usepackage{color}  

\usepackage{bm}
\usepackage{multirow}
\usepackage{bbding}

\journal{Pattern Recognition}

\begin{document}

\begin{frontmatter}



\title{MBQuant: A Novel Multi-Branch Topology Method for Arbitrary Bit-width Network Quantization}


\author[1,2]{Yunshan Zhong}
\author[2,3]{Yuyao Zhou}
\author[2,3]{Fei Chao}
\author[1,2,3,4]{Rongrong Ji}

\affiliation[1]{organization={Institute of Artificial Intelligence, Xiamen University},                 city={Xiamen},
            state={Fujian},
            country={China}}
\affiliation[2]{organization={Key Laboratory of Multimedia Trusted Perception and Efficient Computing, Ministry of Education of China, Xiamen University}, 
        city={Xiamen},
        state={Fujian},
        country={China}}
\affiliation[3]{organization={Department of Artificial Intelligence, School of Informatics, Xiamen University},
            city={Xiamen},
            state={Fujian},
            country={China}}
\affiliation[4]{organization={Peng Cheng Laboratory},             
            city={Shenzhen},
            state={Guangdong},
            country={China}}

\begin{abstract}
Arbitrary bit-width network quantization has received significant attention due to its high adaptability to various bit-width requirements during runtime. 
However, in this paper, we investigate existing methods and observe a significant accumulation of quantization errors caused by switching weight and activations bit-widths, leading to limited performance.
To address this issue, we propose MBQuant, a novel method that utilizes a multi-branch topology for arbitrary bit-width quantization. MBQuant duplicates the network body into multiple independent branches, where the weights of each branch are quantized to a fixed 2-bit and the activations remain in the input bit-width. The computation of a desired bit-width is completed by selecting an appropriate number of branches that satisfy the original computational constraint. By fixing the weight bit-width, this approach substantially reduces quantization errors caused by switching weight bit-widths. Additionally, we introduce an amortization branch selection strategy to distribute quantization errors caused by switching activation bit-widths among branches to improve performance.
Finally, we adopt an in-place distillation strategy that facilitates guidance between branches to further enhance MBQuant's performance.
Extensive experiments demonstrate that MBQuant achieves significant performance gains compared to existing arbitrary bit-width quantization methods. Code is made publicly available at \url{https://github.com/zysxmu/MBQuant}.
\end{abstract}



\begin{keyword}
Network quantization, Quantization-aware training, Arbitrary bit-width, Multi-branch topology


\end{keyword}

\end{frontmatter}



\section{Introduction}

By converting the full-precision weights and activations in deep neural networks (DNNs) into lower-bit formats, the quantization technique has become one of the most predominant methods to show impressive capabilities of compressing DNNs in recent years.
However, most network quantization methods~\cite{gong2019differentiable,CHEN2023109527,yang2021fracbits} are designed with a fixed bit-width, limiting their scalability and adaptability to various computational resources in real-world applications.
Therefore, recent work has focused on training the quantized neural networks (QNNs) of multiple precisions for adapting to a wide range of bit-width requirements in runtime, referred to as arbitrary bit-width QNNs~\cite{yu2021any,jin2020adabits,ijcai2022MultiQuant,bulat2021bit,tang2022arbitrary}.

\begin{figure*}[htbp]
\centering
\hspace{-2cm}
\begin{subfigure}{0.38\linewidth}
\centering
\includegraphics[height=\linewidth]{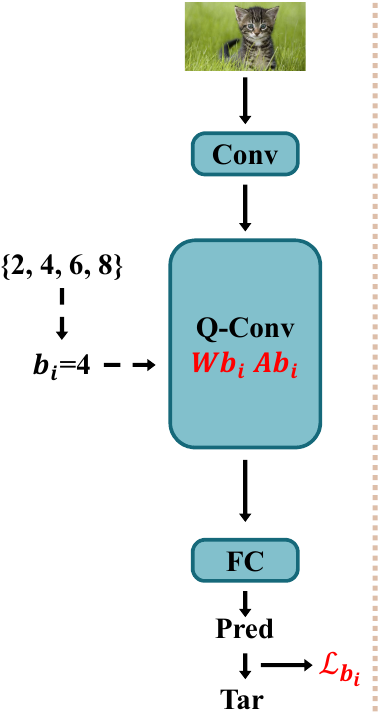}
\subcaption[]{}
\label{fig:framework-baseline}
\end{subfigure}
\begin{subfigure}{0.385\linewidth}
\centering
\includegraphics[height=\linewidth]{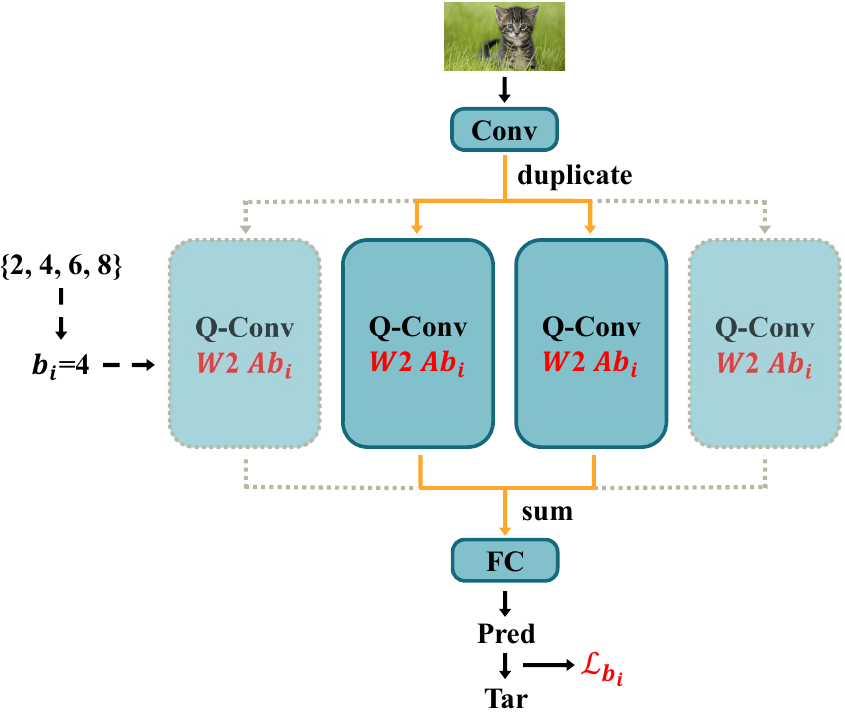}
\subcaption[]{}
\label{fig:framework-our}
\end{subfigure}
\caption{Illustration of the framework of (a) previous methods~\cite{yu2021any,jin2020adabits}. (b) our MBQuant.}
\label{fig:framework}
\end{figure*}

Typically, as illustrated in Fig.\,\ref{fig:framework}a, these methods train the arbitrary bit-width QNNs by optimizing the loss of all bit-width candidates for weights and activations~\cite{jin2020adabits,yu2021any}. 
Specifically, throughout the training process, the bit-width is cyclically chosen from the candidate bit-widths. The weights and activations of the network are then quantized to the selected bit-width. Subsequently, the quantized network undergoes forward and backward propagation to update weights.
Previous methods improve the performance by utilizing mix-precision~\cite{ijcai2022MultiQuant,bulat2021bit}, training strategy~\cite{xu2023eq}, dynamic inference~\cite{tang2022arbitrary}, and so on.
%
%
%
Despite these early efforts, in this paper, we reveal that current methods for arbitrary bit-width QNNs struggle more with quantization errors, making their performance unsatisfactory.


Quantization error, one of the most practical difficulties in QNNs~\cite{ACIQ,zhou2018adaptive,XIE2024110172}, stems from the inherent uncertainty in digitizing an analog value as a result of the finite resolution of the conversion process. It depends typically on the number of bits in the converter, along with its error, noise, and nonlinearities~\cite{web-quanterror,MA2019156}.
In the case of fixed bit-width QNNs, the network only needs to account for quantization errors specific to one bit-width, which generally results in acceptable performance.
However, arbitrary bit-width QNNs encounter more quantization errors due to the switching of bit-widths to accommodate them to different bit-width candidates, as detailed in Sec.\,\ref{sec:quant-error}.
Specifically, during the training of arbitrary bit-width QNNs, weights are quantized to different bit-widths, wherein each bit-width imposes its quantization error on weights.
Thereby the weights have to accommodate the quantization errors from weight quantization of all bit-width candidates, resulting in what we refer to as \textit{quantization error from switching weight bit-widths}. 
Moreover, as pointed out in Wei \emph{et al}.~\cite{weiqdrop}, activation quantization is equal to injecting quantization error to network weights.
During the training, activations are also quantized to different bit-widths, wherein each bit-width leads to a transplanted quantization error to weights.
As a result, network weights have to adapt the transplanted quantization error from activation quantization of all bit-width candidates, which we refer to as \textit{quantization error from switching activations bit-widths}.
These quantization errors pose a significant challenge for arbitrary bit-width QNNs, making it difficult for them to converge to an optimal state and resulting in limited performance.

In this paper, we propose MBQuant, to address significant quantization error in arbitrary bit-width QNNs from a novel perspective of multi-branch topology~\cite{yuslimmable,mishrawrpn,hu2022palquant}.
Specifically, our MBQuant duplicates the network into multiple independent branches and quantizes each branch's weights to the 2-bit format in this paper. Activations, on the contrary, remain at the input bit-width. Then, the computation of a desired bit-width $b_i$ is achieved by combining the output of $\frac{b_i}{2}$ independent branches, where the number of branches is selected to satisfy the original computational constraint.
As illustrated in Fig.\,\ref{fig:framework}b, an 8-bit QNN with 8-bit weights and 8-bit activations is modeled as four branches, each equipped with 2-bit weights and 8-bit activations. A 4-bit QNN is modeled as two branches with each having 2-bit weights and 4-bit activations.
By exploiting the multi-branch topology, weights are fixed to 2-bit, thereby each branch only needs to deal with the quantization error of 2-bit quantization, avoiding handling quantization error from switching weight bit-widths.
In contrast, previous methods only have one branch and the weights have to bear errors from all candidate bit-widths.
Moreover, since each branch only needs to be stored in 2-bit, MBQuant yields low-cost storage.

Then, we introduce an amortization branch selection strategy to amortize quantization error from switching activations bit-widths into branches, in which the selection of branch combination is dispersed. 
Fig.\,\ref{fig:selection} provides an example of the 2-, 4-, 6-, and 8-bit settings. As shown in Fig.\,\ref{fig:selection}a, a simple strategy is the serial branch selection strategy, where the first branch is involved in all bit-widths, making the first branch bear quantization error from 2-, 4-, 6-, and 8-bit activations. As a result, the first branch bears the quantization error from activation quantization too much, leading to a limited performance.
To solve this, as presented in Fig.\,\ref{fig:selection}b, we propose the amortization branch selection strategy, where the selection of branches is dispersed. This strategy only includes the first branch in 2- and 8-bit quantization. As a result, the first branch only needs to adapt the error from 2- and 8-bit activation quantization, leading to a better performance.
Finally, we adopt the in-place distillation~\cite{yu2021any,yuslimmable,yu2020bignas} strategy to facilitate the guidance between branches to further improve MBQuant. Specifically, for the largest bit-width, its optimization objective is the cross-entropy loss. For lower bit-widths, the optimization objectives comprise the cross-entropy loss and the distillation loss from the largest bit-width.

Extensive experiments demonstrate that MBQuant achieves significant performance gains compared to existing arbitrary bit-width quantization methods. For example, MBQuant improves the average accuracy of ResNet-34 by 1.69\% on 2-, 3-, and 4-bit settings, and by 2.66\% on 2-, 4-, 6-, and 8-bit settings.

\section{Related Work}
\label{sec:related}

\subsection{Network Quantization}

Network quantization has been a prominent research topic in the model compression community for a long time. 
Most existing studies can be generally grouped into three parts: quantization-aware training (QAT), post-training quantization (PTQ), and zero-shot quantization (ZSQ).
QAT methods aim to recover the performance of QNNs on the premise of accessing the complete training dataset. They usually focus on quantizer designing~\cite{LIN2023109556,LI2020107409,LSQ}, differentiable quantization~\cite{gong2019differentiable}, regularization~\cite{lee2021cluster,han2021improving} or mix-precision quantization~\cite{CHU2021107647}, \emph{etc}.
PTQ methods, on the other hand, are restricted to a small amount of the training set. Most existing methods attempt to alleviate the accuracy deterioration by designing sophisticated quantizers~\cite{fang2020post} or updating network parameters~\cite{weiqdrop,kim2021distance}.
ZSQ methods instead accomplish network quantization without accessing any real data, which generally synthesize fake images to update quantized network~\cite{choi2021qimera,LI2024110444,zhong2022intraq,CHEN2023109780}.
However, these methods are dedicated to QNNs under the constraint of a specific bit-width and require to re-train QNNs once the constraint changes.

\subsection{Arbitrary Bit-Width Quantization}

Recent development in scalable networks~\cite{yuslimmable,yu2019universally} has led to spurring interest in training arbitrary bit-width QNNs~\cite{YVINEC2024110571,yu2021any,chmiel2020robust,jin2020adabits,bulat2021bit,xu2023eq}. 
\cite{chmiel2020robust} presented a robust regularization method that pushes the network weights to be uniform. \cite{yu2021any} proposed APN where the losses of different bit-widths are summed up to update the model. Besides, switchable batch normalizations are adopted to handle significant distribution gaps between activations of different bit-widths~\cite{yuslimmable}.
AdaBit was proposed by~\cite{jin2020adabits} with similar training paradigms as \cite{yu2021any} while having a different quantizer. To reduce the storage from full-precision to low-bit, AdaBit revises the common round-to-nearest quantizer as the round-to-floor one.
Many techniques are explored to improve the performance of arbitrary bit-width QNNs.
\cite{sun2021improved} designed a collaborative knowledge distillation and a block-swapping method to train the network.
%
%
Bit-Mixer~\cite{bulat2021bit} and MultiQuant~\cite{ijcai2022MultiQuant} adopt mix-precision settings. Specifically, Bit-Mixer~\cite{bulat2021bit} introduces a complicated three-stage training strategy for training the networks. MultiQuant~\cite{ijcai2022MultiQuant} involves a Monte Carlo sampling-based method to select the optimal layer-wise bit-width setting.
ABN~\cite{tang2022arbitrary} further introduces an adaptive inference approach that utilizes a reinforcement learning-based controller to select the optimal layer-wise bit-width for each input sample. Therefore, ABN uses a sample-wise mix-precision inference.
EQ-Net~\cite{xu2023eq} proposes a one-shot network quantization method that adopts various bit-width, quantization granularity, and quantizer symmetry for different layers. 
\cite{liu2022instance} enabled layer-wise mix-precision, sophisticated training recipes, and finer granularity.
Despite these advances, these methods only train one-branch quantized network, thus suffer from significant quantization error from the switching of weights and activations bit-widths and thus are limited in performance.

\section{Method}

\subsection{Preliminaries}
\label{sec:preliminaries}

Following the settings of~\cite{hu2022palquant}, we use a uniform quantizer with trainable clipping parameters to implement network quantization. 
Given a full-precison value $\bm{x}$, either representing weights or activations, the quantizer first normalizes $\bm{x}$ to a range $[0, 1]$ by the following equation:
\begin{equation}
\bm{x}_n = clip(\frac{\bm{x}-l}{u-l}, 0, 1),
\label{quantizer1}
\end{equation}
where $l$ is the trainable upper bound, $u$ is the trainable lower bound, $clip(\cdot, 0, 1)$ is the clipping function. Then the normalized value $\bm{x}_n$ is quantized to obtain the integer $\bm{q}$:
\begin{equation}
\bm{q} = \lfloor (2^b - 1) \times \bm{x}_n)\rceil,
\label{quantizer2}
\end{equation}
where $\lfloor \cdot \rceil$ rounds its input to the nearest integer. The corresponding de-quantized value $\bar{\bm{x}}$ can be calculated as:
\begin{equation}
\begin{aligned}
\bar{\bm{x}}  =
    \begin{cases}
      2 \times \big (\frac{\bm{q}}{2^b-1} - 0.5 \big), & \bm{x} \in weights. \\
    \frac{\bm{q}}{2^b-1}, & \bm{x} \in activations. 
    \end{cases}
\end{aligned}
\end{equation}

For activations and weights, we both use layer-wise quantizer. We do not quantize the first convolutional layer and the last full-connected layer to follow~\cite{yu2021any}.

\subsection{Quantization Error in Arbitrary Bit-Width QNNs}
\label{sec:quant-error}

As illustrated in Fig.\,\ref{fig:framework}a, to accommodate QNNs to different bit-widths, previous methods~\cite{jin2020adabits,yu2021any,ijcai2022MultiQuant,tang2022arbitrary} switch the bit-width of network weighs and activations continuously and perform forward and backward process. Unfortunately, current arbitrary bit-width QNNs usually suffer from a significant quantization error as we detail in the following.

\textbf{Quantization Error from Switching Weight Bit-widths}.
Assuming the full-precision weight $\bm{w}$ follows a normal distribution with probability density function (PDF) as $f(\bm{w}) \sim \mathcal N(0, 1)$~\cite{ACIQ}, the bit-width is $b$, each full-precision $\bm{w}$ is rounded to the midpoint of its quantization bin $\bar{\bm{w}}$, the upper bound of weights is $u^w$, the lower bound of weights is $l^w$, and $l^w=-u^w$, the quantization error caused by weight quantization can be defined as:
\begin{equation}
\begin{split}
\text{MSQE}_b & = E[(\bm{w} - \bar{\bm{w}})^2]  \approx \underbrace{\frac{(u^w)^2}{3 \times 2^{2b}}}_{Quantization \, Noise} + \underbrace{2 \times \int_{u^w}^{\infty} f(\bm{w}) (\bm{w} - u^w) d\bm{w}}_{Clipping \, Noise}.
\end{split}
\label{MSQE}
\end{equation}

A systematic derivation can be found in~\cite{ACIQ,chmiel2020robust}.
%
A small upper bound parameter $u^w$ causes less quantization noise but more clipping noise.
Note that Eq.\,(\ref{MSQE}) varies with alterations in the weight distribution, lower bound, and upper bound, but it does not affect the conclusion that weights suffer from quantization noise.

For single bit-width QNNs, they only need to adjust the weights to accommodate the error from weight quantization of a specific bit-width $b_i$, \emph{i.e.}, $\text{MSQE}_{b_i}$. In contrast, for arbitrary bit-width QNNs, they have to adjust the weights to accommodate errors from weight quantization of all possible bit-width candidates as:
\begin{equation}
\begin{split}
Error_{w} = \sum_{b_i} \text{MSQE}_{b_i}.
\end{split}
\label{Eq:WE}
\end{equation}

We denote Eq.\,\ref{Eq:WE} as the \textit{quantization error from switching weight bit-widths}.
Given the example shown in Fig.\,\ref{fig:framework}a, for training arbitrary bit-width QNNs that support 2-, 4-, 6-, and 8-bit quantization, previous methods typically train the weights to accommodate quantization error from 2-, 4-, 6-, and 8-bit weight quantization, \emph{i.e.}, $\text{MSQE}_{2} + \text{MSQE}_{4} + \text{MSQE}_{6} + \text{MSQE}_{8}$.
Compared with single bit-width QNNs, arbitrary bit-width QNNs suffer from more quantization errors from weight quantization, making it a challenge to converge in an optimal state.

\textbf{Quantization Error from Switching Activation Bit-widths}.
As analyzed in \cite{weiqdrop}, the quantization of activations is equivalent to injecting quantization error into the network weights. Assuming the activations are quantized into $b$ bit, the transplanted quantization error can be represented by:
\begin{equation}
\bm{w} \odot \bar{\bm{a}}_b = \bm{w} \odot \Big(\bm{a}\big(\bm{1}+\bm{n}_b^a(\bm{I})\big)\Big) \approx \Big(\bm{w}\big(\bm{1}+\bm{n}_b^w(\bm{I})\big)\Big) \odot \bm{a},
\label{noise-trans}
\end{equation}
where $\odot$ denotes the convolutional operation, $\bm{n}_b^a(\bm{I})$ represents the activation quantization error that is dependent on the distribution of input data $\bm{I}$ 
and $\bm{n}_b^w(\bm{I})$ represents the transplanted quantization error on weight from the activation quantization.
A systematic derivation can be found in \cite{weiqdrop}.

For single bit-width QNNs, they only need to adjust the weights to accommodate quantization error transplanted from activation quantization of a specific bit-width $b_i$, \emph{i.e.}, $\bm{n}^w_{b_i}(\bm{I})$. In contrast, for arbitrary bit-width QNNs, the activations bit-width also switches among candidate bit-widths during the training period, resulting in accumulated transplanted quantization error on network weights as:
\begin{equation}
\begin{split}
Error_{a} = \sum_{b_i} \bm{n}^w_{b_i}(\bm{I}).
\end{split}
\label{Eq:AE}
\end{equation}

We denote Eq.\,\ref{Eq:AE} as \textit{quantization error from switching activations bit-widths}.
As the example in Fig.\,\ref{fig:framework}a, for training arbitrary bit-width QNNs that support 2-, 4-, 6-, and 8-bit quantization, previous methods typically train the weights to accommodate quantization error transplanted from activation quantization of 2-, 4-, 6-, and 8-bit quantization, \emph{i.e.}, $\bm{n}^w_{2}(\bm{I}) + \bm{n}^w_{4}(\bm{I}) + \bm{n}^w_{6}(\bm{I}) + \bm{n}^w_{8}(\bm{I})$.
Compared with single bit-width QNNs, arbitrary bit-width QNNs suffer from more quantization errors from activation quantization.
As a result, the error from activation quantization further exacerbates the convergence of network weights, especially considering that the weights have already undergone quantization error from switching weight bit-widths.

In summary, for training an arbitrary bit-width QNN, the quantization error consists of the error from switching weight bit-widths and the error from switching activation bit-widths:
\begin{equation}
\begin{split}
Error = \sum_{b_i} \text{MSQE}_{b_i} + \bm{n}^w_{b_i}(\bm{I}).
\end{split}
\end{equation}

As for a single bit-width QNN, the quantization error is:
\begin{equation}
\begin{split}
Error = \text{MSQE}_{b_i} + \bm{n}^w_{b_i}(\bm{I}).
\end{split}
\end{equation}

It can be observed that the arbitrary bit-width QNN suffers from more quantization error than the single bit-width QNN, thereby usually leading to limited performance.
Unfortunately, previous works typically overlook this large error, resulting in unsatisfactory performance. 
In the following, we first present a multi-branch topology~\cite{yuslimmable,mishrawrpn,hu2022palquant} that duplicates a network into multiple fixed bit-width branches to attenuate quantization error from switching
weight bit-widths, and then introduce an amortization branch selection strategy to amortize quantization error from switching activation bit-widths.

\subsection{MBQuant}

Towards mitigating quantization error from switching
weight bit-widths in arbitrary bit-width QNNs, we first present MBQuant, a novel quantization method from the perspective of multi-branch topology~\cite{yuslimmable,mishrawrpn,hu2022palquant}.
\begin{wraptable}{r}{0.5\textwidth} 
\footnotesize
\caption{Quantization error of the layer1.conv1 of ResNet-20 on the CIFAR-100 with 2-, 4-, 6-, and 8-bit settings. Results of APN~\cite{yu2021any}, AdaBit~\cite{jin2020adabits}, and our MBQuant are reported.}
\centering
\resizebox{\linewidth}{!}{ 
\begin{tabular}{c|ccc}
\bottomrule[1.25pt]
    Epoch              & APN                    & AdaBit                   &  MBQuant (Ours)  \\ \hline \hline
80  & 0.4449  & 0.4415 & 0.2821 \\ \hline
120 & 0.4357 &  0.4109  &   0.2728  \\ \hline
160 & 0.4322 & 0.3968  & 0.2721 \\ \bottomrule[1pt]
\end{tabular}
}
\label{tab:quant-errors}
\end{wraptable}
Fig.\,\ref{fig:framework}b illustrates an example of our multi-branch topology. Considering a bit-width candidate set $\{b_i\}_{i=1}^M$, we make $\frac{\max(b_i)}{2}$ exact copies of the network body, and weights of each branch are quantized to 2-bit.
Then, for the $b_i$-bit quantization, we retain the activation in the $b_i$-bit state and choose a total of $b_i/2$ branches as an alternative option to the $b_i$-bit weights. Here, we select 2-bit since the 2-bit is the smallest bit-width for the most current arbitrary bit-width QNNs methods~\cite{jin2020adabits,bulat2021bit,tang2022arbitrary}.
For an incoming input $\bm{I}$, we obtain the output of the first full-precision convolutional layer, the result of which is subsequently duplicated by $b_i/2$ times. The duplicated outputs are regarded as the input of the selected branches to perform forward pass independently.
Finally, we sum up all outputs from these branches to yield the input of the last fully-connected layer.
The entire process for $b_i$-bit quantization is formulated as:
\begin{equation}
    output_{b_i} = \mathbf{FC}\Big(\sum_{j \in \mathcal{P}}  branch_j\big(\mathbf{Conv_1}(\bm{I})\big)\Big),
\end{equation}
where $\mathcal{P}$ is the index set of the selected branches and $\|\mathcal{P}\| = \frac{b_i}{2}$.
For example, when attempting to execute an 8-bit QNN, four branches are adopted, each of which processes 2-bit weights and 8-bit activations. 
Similarly, a 4-bit QNN can be achieved by selecting two branches, each of which processes 2-bit weights and 4-bit activations.

It is worth noting that MBQunt presents comparable computational costs to previous methods that adopt a single branch.
To be specific, the computation involved in executing $\frac{b_i}{2}$ branches, each with 2-bit weight and $b_i$ activations, can be denoted as $\frac{b_i}{2} \times \bar{\bm{w}}_{2} \odot \bar{\bm{a}}_{b_i}$. Notably, this is equivalent to $\bar{\bm{w}}_{b_i} \odot \bar{\bm{a}}_{b_i}$, which corresponds to the computation of a single branch with both $b_i$ weights and activations.
Thereby the extra computation is only caused by the summation of the outputs from branches, which, however, is negligible.
For example, when executing 8-bit ResNet-18 on ImageNet, the Bit-Operations (BOPs) of previous methods is 229.76G. In contrast, utilizing MBQuant increases the BOPs slightly to 229.81G, incurring a negligible additional cost of only 0.02\%.

To handle an uneven $b_i$, the multi-branch topology consists of $\lfloor b_i/2 \rfloor$ branches as well as an extra half branch with its channel number by halves.
The input of the half branch is sequentially chosen from the output of the first convolutional layer, based on the input channel amount of the second branch.
Also, the channels of the last convolutional layer of the half branch remain as the original to avoid dimension mismatch with the fully-connected layer.
Such designation provides a residual learning mode where the newly added branch learns the residual between the results of previous branches and the ground truth~\cite{he2016deep,yuslimmable}.
By deploying our multi-branch topology, the weights of each branch within MBQuant are fixed to 2-bit, thus the need for switching different weight bit-widths in previous arbitrary bit-width QNNs methods is avoided. Specifically, each branch only needs to handle the error from 2-bit weight quantization, \emph{i.e.}, $\text{MSQE}_{\text{2}}$. In contrast, previous methods only have one branch and the weights have to bear errors from all candidate bit-widths, \emph{i.e.}, $\sum_{b_i} \text{MSQE}_{b_i}$, disturbing its convergence and resulting suboptimal performance. 
Thus, in our MBQuant, the quantization error from switching weights bit-widths is effectively eliminated, making it easier to converge to an optimal state. The quantitative results of Table\,\ref{tab:quant-errors} also present a consistent conclusion that MBQuant provides a lower weight quantization error than other methods.

%
%
%

We note an earlier study~\cite{hu2022palquant} also considers a similar multi-branch setting. Our MBQuant fundamentally differs in application area and motivation. 
First, MBQuant aims for arbitrary bit-width QNNs while \cite{hu2022palquant} targets at a specific bit-width.
Second, MBQuant starts from the analysis of the specific quantization error in current arbitrary bit-width methods while \cite{hu2022palquant} started from the deployment of high-bit models on low-bit accelerators.
\begin{figure}[!t]
\centering
\includegraphics[width=0.7\linewidth]{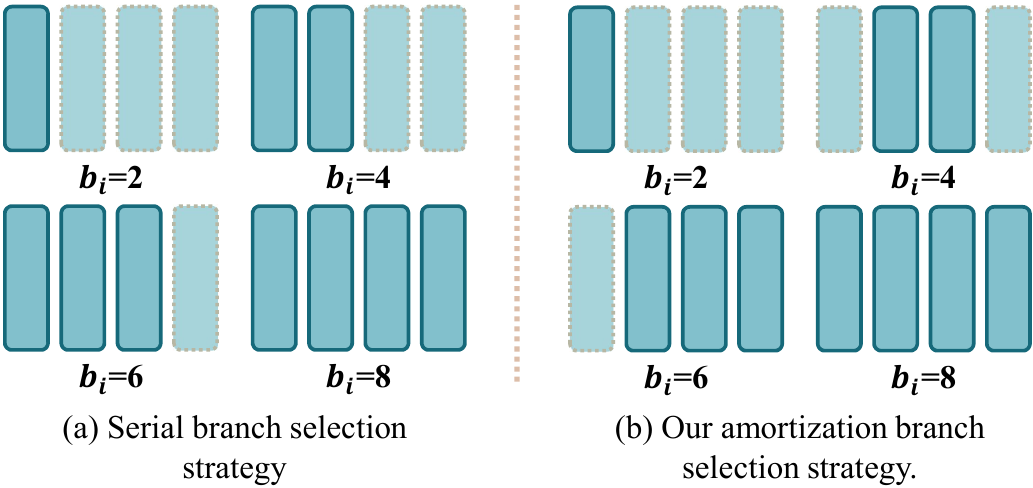}
\caption{Illustration of the (a) serial branch selection strategy. (b) our amortization branch selection strategy.}
\label{fig:selection}
\end{figure}

\subsection{Amortization Branch Selection Strategy}

Though accumulated weight quantization error is relieved by our multi-branch topology, the error from quantizing activations is unavoidable.
We then propose an amortization branch selection strategy to amortize quantization error from switching activations bit-widths into branches.

In particular, given the topology employed by MBQuant, a straightforward strategy is to serially select the branches. The example of 2-, 4-, 6-, and 8-bit is shown in Fig.\,\ref{fig:selection}a. It can be found that $\mathcal{P}=\{1\}$, $\mathcal{P}=\{1, 2\}$, $\mathcal{P}=\{1, 2, 3\}$, and $\mathcal{P}=\{1, 2, 3, 4\}$ for 2-, 4-, 6-, and 8-bit, respectively.
However, this strategy selects the first branch in all candidate bit-widths, making it bear all errors from 2-, 4-, 6-, and 8-bit activation quantization: $\bm{n}^w_{2}(\bm{I})+\bm{n}^w_{4}(\bm{I})+\bm{n}^w_{6}(\bm{I})+\bm{n}^w_{8}(\bm{I})$.
As a result, the first branch is hard to converge to the optimal state, leading the performance degradation for the first branch.

To handle this issue, in our amortization branch selection strategy, the selection of branches is dispersed to avoid the error being concentrated as illustrated in Fig.\,\ref{fig:selection}b.
Specifically, $\mathcal{P}=\{1\}$, $\mathcal{P}=\{2, 3\}$, $\mathcal{P}=\{2, 3, 4\}$, and $\mathcal{P}=\{1, 2, 3, 4\}$ for 2-, 4-, 6-, and 8-bit, respectively. Therefore, the first branch is only selected in the 2- and 8-bit cases. In contrast, the second and third branch is used in the 4-, 6-, and 8-bit cases, and the last branch is selected in 6- and 8-bit cases. Correspondingly, the first branch now only has to bear the error from 2- and 8-bit activation quantization: $\bm{n}^w_{2}(\bm{I})+\bm{n}^w_{8}(\bm{I})$.
Compared with the serial branch selection strategy, the first branch in our strategy bears fewer errors from activation quantization.
\begin{algorithm*}[!h]
\footnotesize
\caption{Overall process.}\label{alg:alg1} 
\begin{algorithmic}[1]  
    \REQUIRE Candidate bit-widths list $\mathcal{S}=\{b_i\}_{i=1}^{K}$, training set $\mathcal{D}_t$
    \STATE Initialize the model $\mathcal{F}$.
    \WHILE{not end}
        \STATE Sample data batch $(I, y)$ from train set $\mathcal{D}_t$.
        \FOR{$i = K$ \TO $1$}
            \STATE Set bit-width: $\mathcal{F} \gets b_i$.
            \STATE Select branch $\mathcal{P}$ according to $b_i$ and the amortization branch selection strategy.
            \STATE Get prediction: $Pred_{b_i} \gets \mathcal{F}_{b_i}(I)$.
            \IF{$i==K$}
                \STATE ${\cal L}_{b_i} = CE\big(Pred_{b_i}, y\big)$.
            \ELSE
                \STATE ${\cal L}_{b_i} = CE\big(Pred_{b_i}, y\big) + KD\big(Pred_{b_i}, Pred_{b_K}\big)$.
            \ENDIF
            \STATE ${\cal L} = {\cal L} + {\cal L}_{b_i}$
            \STATE Back-propagate to update network parameters.
        \ENDFOR
    \ENDWHILE
\end{algorithmic}
\end{algorithm*}

It is worth noting that despite the total quantization error incurred by activation quantization not being reduced, the unbalanced allocation of the error is mitigated. As a result, the performance of the first branch is improved.

\subsection{In-place Distillation}

To facilitate the guidance between branches to further improve MBQuant, we adopt the common in-place distillation strategy~\cite{yu2021any,yuslimmable,yu2020bignas}. Specifically, for the largest bit-width $b_K$, its optimization objective is the cross entropy between prediction and the one-hot label. For other bit-widths, their optimization objective is the combination of the cross entropy loss and the knowledge distillation loss from the largest bit-width. The loss function is defined as:
\begin{equation}
\begin{split}
    \begin{cases}
     & if \, i == K:  \quad     {\cal L}_{b_i} = CE\big(Pred_{b_i}, y\big),
    \\
     & if \, i \neq K: \quad    {\cal L}_{b_i}  = CE\big(Pred_{b_i}, y\big)  +  KD\big(Pred_{b_i}, Pred_{b_K}\big),
    \end{cases}
\end{split}
\end{equation}
where $K$ is the index of largest bit-width, $CE(\cdot, \cdot)$ represents the cross entropy loss, $KD(\cdot, \cdot)$ represents the knowledge distillation loss that is the soft cross entropy~\cite{yu2021any}, $y$ is the one-hot label.

\subsection{Algorithm Process}

Our algorithmic process utilized in this paper is presented comprehensively in Alg.\,\ref{alg:alg1}.
By incorporating all the proposed techniques, our MBQuant effectively mitigates quantization error inherent in existing arbitrary bit-width QNNs. Compared to existing methods, MBQuant exhibits significant performance gains.


\begin{table*}[!ht]
\caption{Results of ResNet-18 and ResNet-34 on ImageNet dataset. Results of APN and AdaBit are reproduced by their official code. The number within brackets of APN is copied from the original paper for a comprehensive comparison. Results of other methods are obtained from their original paper. ``Ind.'' indicates the individual training. ``Avg.'' indicates the average accuracy over all candidate bit-widths. ``$\dagger$'' indicates using the round-to-floor quantizer. ``$\ddagger$'' indicates using the center quantizer.}
\centering
\resizebox{\linewidth}{!}{
\begin{tabular}{ccccccccccc}
\bottomrule[1.25pt]
\multirow{2}{*}{Networks} & \multirow{2}{*}{\begin{tabular}[c]{@{}c@{}}Bit-Widths \\ Size\end{tabular}} &  \multicolumn{8}{c}{Methods}  \\ \cline{3-10} 
                          &                             & Ind.  & APN  & AdaBit$\dagger$ & AdaBit$\ddagger$  & Bit-Mixer  & ABN & MultiQuant &  MBQuant (Ours)   \\ \hline\hline
\multirow{11}{*}{ResNet-18}  & 4                           &   69.95 & 66.68$_{(67.96)}$ &  67.90 & 68.99 & 69.4 & 68.9 & / &   \textbf{70.49}    \\
                          & 3                           &   68.81    &  66.28 &  67.05 & 68.48 & 68.7 &   68.6 & / &   \textbf{69.56}    \\
                          & 2                           &   65.79   & 65.07$_{(64.19)}$ & 20.73  & 64.86   & 65.6  & 65.5 & /  &  \textbf{67.31}    \\ 
                          & Avg.                           & 68.18   &  66.01  &  51.89 & 67.44 & 67.9 &  67.67 & / &  
 \textbf{69.12} \\ 
                          & Size (MB)                           &    - & 46.84   &   7.78   & 7.78 & 7.78 & 7.78 & / &    7.85    \\ \cline{2-10} 
                          & 8                           & 70.48      & 66.58$_{(68.04)}$   &  69.10    & 69.82 & / & / & 70.17 &   \textbf{71.07}     \\
                          & 6                           & 70.34  & 66.56  & 69.07  & 69.81 & / &  /  & 69.99 &
 \textbf{70.80}    \\
                          & 4                           & 69.95    & 66.45$_{(67.96)}$ & 68.45  & 69.51 & / & / & 69.68 &   \textbf{70.13}    \\
                          & 2                           & 65.79   & 64.61$_{(64.19)}$ & 25.07 & 63.68  & /  & / & 66.56 &   \textbf{66.84}    \\ 
                          & Avg.                           & 69.14 &  66.07  & 57.91 & 68.21 & /  & / & 69.10 &   \textbf{69.71}      \\ 
                          & Size (MB)                           &  -   &  46.87   &   13.40  &  13.40  & -  & - & 46.87 &  13.63  \\ \hline
\multirow{11}{*}{ResNet-34}  & 4                           &  74.68  & 70.13  & 57.10 &  72.88 & 73.0 & 73.5 & / &   \textbf{74.90}    \\
                          & 3                           &  72.56  & 69.99 & 53.76 & 72.48   & 72.6 & 73.0  & / &   \textbf{74.45}   \\
                          & 2                           & 71.57   & 68.91  & 3.59 &  69.37  &  70.1 &  70.3 & / &   \textbf{72.53}     \\ 
                          & Avg.                           &    72.93   & 69.68  &   38.15  & 71.58 & 71.90 &  72.27 & / &   \textbf{73.96}  \\ 
                          & Size  (MB)                          & -  & 87.33 &  12.92    & 12.92  & 12.92 & 12.92 & - &   13.04  \\ \cline{2-10} 
                          & 8                           & 75.12      & 70.26   & 61.37    & 73.36 & / & / & / &   \textbf{75.73}     \\
                          & 6                           & 74.96   & 70.43 & 61.25 & 73.32  & / & / & / &   \textbf{75.35}   \\
                          & 4                           & 74.68   & 70.36 & 59.81    &  73.16 & / & / & / &   \textbf{75.01}   \\
                          & 2                           & 71.57 & 69.03 &   1.14    & 68.41  & / & / & / &   \textbf{72.74}     \\ 
                          & Avg.                           &  74.08 & 70.02 &  45.89 &  72.05 & / & / & / &   \textbf{74.71} \\ 
                          & Size (MB)                           &   -   & 87.40   &   23.62     & 23.62 & - &  -& - &    24.03   \\ \bottomrule[1.0pt]   
\end{tabular}
}
\label{tab:ImageNet-resnet}
\end{table*}

\begin{table*}[ht]
\footnotesize
\caption{Comparisons with Bit-Mixer~\cite{bulat2021bit}, ABD~\cite{tang2022arbitrary}, and EQ-Net~\cite{xu2023eq}. ``$\ddagger$'' indicates the mix-precision setting.}
\centering
\begin{tabular}{ccccccc}
\bottomrule[1.25pt]
\multirow{2}{*}{Networks} & \multirow{2}{*}{\begin{tabular}[c]{@{}c@{}}Bit-Widths \\ Size\end{tabular}} &  \multicolumn{5}{c}{Methods}  \\ \cline{3-7} 
                          &        & Ind.                 & Bit-Mixer$\ddagger$ & ABN$\ddagger$ & EQ-Net$\ddagger$ &  MBQuant (Ours)   \\ \hline\hline
\multirow{5}{*}{ResNet-18}  & 4         & 69.95        &  69.2 &   69.8 & / &\textbf{70.49}    \\
                          & 3        & 68.81        &  68.6 &  69.0 & 69.3 &\textbf{69.56}    \\
                          & 2        & 65.79           &  64.4  &   66.2 & 65.9 &\textbf{67.31}    \\ 
                          & Avg.    & 68.18            & 67.4  &   68.33 & / & \textbf{69.12} \\ 
                          & Storage size   & -              &  7.78 &   46.84 & 46.84 & 7.85     \\ \bottomrule[.75pt]
\multirow{5}{*}{ResNet-34}  & 4      & 74.68         & 72.9 &   74.0 & / & \textbf{74.90}    \\
                          & 3       & 72.56         & 72.5 &      73.3 & / & \textbf{74.45}   \\
                          & 2      & 71.57            &  69.6 &   71.7 & / & \textbf{72.53}     \\ 
                          & Avg.      & 72.93        &  71.67     & 73.0 & / & \textbf{73.96}  \\ 
                          & Storage size   & -           &  12.92    & 46.84 & - & 13.04     \\ \bottomrule[1.0pt]                          
\end{tabular}
\label{tab:bitmix}
\end{table*}

\begin{table*}[ht]
\footnotesize
\caption{Comparisons with MultiQuant~\cite{ijcai2022MultiQuant}. ``$\ddagger$'' indicates the mix-precision setting. ``FP'' denotes the network is stored with full-precision.}
\centering
\begin{tabular}{ccccc}
\bottomrule[1.25pt]
\multirow{2}{*}{Networks} & \multirow{2}{*}{\begin{tabular}[c]{@{}c@{}}Bit-Widths \\ Size\end{tabular}} &  \multicolumn{3}{c}{Methods}  \\ \cline{3-5} 
                          &                & Ind.          & MultiQuant$\ddagger$  & MBQuant (Ours)   \\ \hline\hline
\multirow{5}{*}{ResNet-18}  & 8         & 70.48          & / &   \textbf{71.07}     \\
                          & 6          & 70.34       & 70.62
  & \textbf{70.80}    \\
                          & 4         & 69.95       &  69.66
  &  \textbf{70.13}    \\ 
                          & 2          & 65.79      &   /  &  \textbf{66.84}    \\ 
                          & Avg.         & 69.14     &  / &  \textbf{69.71} \\ 
                          & Storage size    &   -       &  46.84 &   7.85     \\ \bottomrule[1.0pt]                          
\end{tabular}
\label{tab:multi}
\end{table*}

\section{Experimentation}

\subsection{Setups}

\textbf{Datasets and Networks}.
The experimental evaluations are conducted using the widely-used CIFAR-100~\cite{cifar} and ImageNet~\cite{russakovsky2015imagenet} datasets.
We quantize ResNet-20~\cite{he2016deep} for CIFAR-100, and ResNet-18, ResNet-34, and MobileNetV1~\cite{howard2017mobilenets} for ImageNet.
The candidate bit-widths for the experiments include two settings: 2-, 3-, and 4-bit, and 2-, 4-, 6-, and 8-bit.

\textbf{Training Settings}.
All experiments are implemented with Pytorch framework. During the training phase, the gradient of the rounding function is approximated by the straight-through estimator(STE)~\cite{QIN2020107281}.
The Adam optimizer~\cite{kingma2014adam} is used for quantization parameters, with an initial learning rate of 1e-4 and a weight decay of 0. For model parameters, the SGD optimizer with momentum set to 0.9 is adopted, and the learning rate and weight decay are set to 0.01 and 1e-4, respectively.
For all models, we load the pre-trained full-precision checkpoint to provide a good initialization.
For CIFAR-100, a batch size of 128 and a training epoch of 200 are used, with a learning rate decay by a factor of 0.1 at 100 epochs and 150 epochs. The data augmentation consists of ``random crop'' and ``random horizontal flip''.
For ImageNet, the batch size is 256 and the training epoch is 90. The learning rate is adjusted by the cosine learning rate decay strategy. Standard data augmentation is used, including ``random resize and crop'', and ``random horizontal flip''.
All experiments are implemented with 4 NVIDIA 3090 GPUs.

\begin{table*}[ht]
\footnotesize
\caption{Results of MobileNetV1 on ImageNet dataset. Results of APN and AdaBit are reproduced by their official code. ``/'' indicates the model is collapsed.}
\centering
\begin{tabular}{ccccccc}
\bottomrule[1.25pt]
\multirow{2}{*}{Networks} & \multirow{2}{*}{\begin{tabular}[c]{@{}c@{}}Bit-Widths \\ Size\end{tabular}} &  \multicolumn{4}{c}{Methods}  \\ \cline{3-7} 
                          &                             & Ind.  & APN  & AdaBit$\dagger$ & AdaBit$\ddagger$  & MBQuant (Ours)   \\ \hline\hline
\multirow{11}{*}{MobileNetV1}  & 4                           &  70.17  &  64.60 & /  & 65.92 &  66.13 \\
                          & 3                           &     67.68  &  62.94 &  / & 64.19 & 64.97 \\
                          & 2                           &  52.94    & 53.57  & /  & 49.93   & \textbf{55.75}    \\ 
                          & Avg.                           &  63.60  & 60.37 &  / & 60.01  & 62.28 \\ 
                          & Size (MB)                           &   -  &  17.10   &  -    & 5.96  & 6.11     \\ \cline{2-7} 
                          & 8                           &  71.83     & 66.36   &    /  &  70.36 & 70.53   \\
                          & 6                           & 71.60  & 65.94  &  / &  70.31  & \textbf{72.03}   \\
                          & 4                           & 70.17    & 65.44  &  / & 69.55 & 70.09     \\
                          & 2                           &  52.94  & 54.43 & / & 50.27   & \textbf{54.46}  \\ 
                          & Avg.                           & 66.64 & 63.04 & / & 65.12  & \textbf{66.78}     \\ 
                          & Size (MB)                           &  -   & 17.19  &    -  & 7.64 & 8.16    \\ \bottomrule[1.0pt]   
\end{tabular}
\label{tab:ImageNet}
\end{table*}

\textbf{Implementation Details}.
Consistent with previous work~\cite{yu2021any}, the first and last layers are not quantized in our experiments, and are shared among branches. Moreover, the batch normalization layer is also not quantized, consistent with prior methods~\cite{jin2020adabits, bulat2021bit}.
Each branch's weights are quantized to 2-bit with a single pair of quantization parameters ($l^w, u^w$). For the activation of each branch, switchable batch normalization and independent quantization parameters ($l^a_i, u^a_i$) are used for each bit-width $b_i$ to address the distribution gap between bit-widths, following the approach of previous studies~\cite{jin2020adabits, bulat2021bit}. 

\textbf{Compared Methods}.
In the main paper, we compared MBQuant with APN~\cite{yu2021any}, AdaBit~\cite{jin2020adabits}, Bit-Mixer~\cite{bulat2021bit}, ABN~\cite{tang2022arbitrary}, ABD~\cite{tang2022arbitrary}, EQ-Net~\cite{xu2023eq}, and MultiQuant~\cite{ijcai2022MultiQuant}. We reproduce APN~\cite{yu2021any} and AdaBit~\cite{jin2020adabits} based on their official open-source code. For a comprehensive comparison, we respectively implement AdaBit with the round-to-floor quantizer proposed in their paper, and with the center quantizer proposed in their GitHub repository. The results of other methods are obtained from their paper.

\subsection{Results on ImageNet}

\subsubsection{ResNet18 and ResNet34}

Our MBQuant adopts the uniform precision setting, \emph{i.e.}, all layers adopt the same bit-width. Thus, in Table\,\ref{tab:ImageNet-resnet}, we first provide comparisons between our MBQuant with the results from other methods that also adopt the uniform precision setting. It can be seen that our MBQuant achieves the best accuracy on ResNet-18 and ResNet-34 when tested on the ImageNet dataset across bit-widths under various settings.

In particular, on 2-, 3-, and 4-bit settings, MBQuant obtains 67.31\%, 69.56\%, and 70.49\% top-1 accuracy for ResNet-18. As a result, MBQuant achieves accuracy gains of 1.71\%, 0.86\%, and  1.09\% for 2-, 3-, and 4-bit, respectively. 
On 2-, 4-, 6-, and 8-bit settings, the accuracy gains of MBQuant are 0.28\%, 0.45\%,  0.81\%, and 0.90\%, respectively. Consequently, the average accuracy is improved by 1.22\% and 0.61\% for these two settings, respectively.
Results of ResNet-34 show similar conclusions that MBQuant outperforms previous methods by a large margin. Specifically, on 2-, 3-, and 4-bit settings, MBQuant presents 72.52\%, 74.45\%, and 74.90\% top-1 accuracy, which corresponding to 2.23\%, 1.45\%, and 1.40\% accuracy improvements, respectively. As a result, the average accuracy is improved by 1.69\%. Also, on 2-, 4-, 6-, and 8-bit settings, MBQuant presents 72.74\%, 75.01\%, 75.35\%, and 75.73\% top-1 accuracy, which corresponding to 3.71\%, 1.85\%, 2.02\%, 2.37\%  accuracy gains. Consequently, MBQuant obtains gains on average accuracy of 2.66\% for this setting.
MBQuant also presents size superiority. On 2-, 3-, and 4-bit settings, MBQuant only respectively occupies 7.78MB and 13.04 MB for ResNet-18 and ResNet-34 while APN occupies 46.84MB and 87.33 MB. 
Notably, MBQuant presents better accuracy than individual training. In particular, for all bit-width, MBQuant demonstrates better accuracy than the individual training for ResNet-18 and ResNet-34.

We further provide comparisons between our MBQuant with other arbitrary bit-width methods that adopt mix-precision settings. To be brief, the compared Bit-Mixer~\cite{bulat2021bit} and MultiQuant~\cite{ijcai2022MultiQuant} utilize mix-precision setting for the input samples in the inference. ABN~\cite{tang2022arbitrary} uses different mix-precision settings dependent on the input sample in the inference. EQ-Net~\cite{xu2023eq} involves once-for-all model training and mixed-precision quantization search.
Note that our MBQuant only adopts uniform precision for all layers within the networks and does not involve sample-wise inference.

We directly use the results reported in their paper despite the hyperparameter configuration, total epochs, and training strategy being very different.
The comparisons between our MBQuant and Bit-Mixer and ABN are presented in Table\,\ref{tab:bitmix}. Note that they use 160 training epochs while MBQuant only uses 90 epochs. 
It can be seen that our MBQuant achieves the best results in all settings. For example, at 2-, 3-, and 4-bit settings, MBQuant respectively achieves 0.69\%, 0.56\%, and 1.11\% improvements on ResNet-18. While on ResNet-34, MBQuant achieves 0.90\%, 1.15\%, and 0.83\% improvements for 2-, 3-, and 4-bit cases, respectively.
As a result, MBQuant presents 0.79\% and 0.96\% average accuracy improvements for ResNet-18 and ResNet-34, respectively.
The comparison between our MBQuant and MultiQuant~\cite{ijcai2022MultiQuant} is presented in Table\,\ref{tab:multi}. It can be seen that our MBQuant exhibits better accuracy than MultiQuant in most bit-widths while still reducing the storage costs by a large margin. Specifically, MBQuant improves the accuracy by 0.28\%, 0.45\%, 0.18\%, and 0.9\% for 2-, 4-, 6-, and 8-bit. Therefore, the average accuracy gains of our MBQuant is 0.61\%. Moreover, the storage size of MBQuant is much lower than MultiQuant.

\subsubsection{MobileNetV1}

Similar to the results of ResNet, MBQuant also improves performance when applied on MobileNetV1. For example, on 2-, 3-, and 4-bit settings, MBQuant obtains 55.75\%, 64.97\%, and 66.13\% top-1 accuracy, corresponding to 2.18\%, 0.78\%, and 0.21\% accuracy improvements, respectively. For 2-, 4-, 6-, and 8-bit settings, MBQuant presents improvements of 4.19\%, 0.54\%, 1.72\%, and 0.17\%, respectively. Correspondingly, the average gains respectively are 1.91\% and 1.66\% for these two settings.
For MobileNetV1, the size of MBQuant is 6.11 MB while APN is 17.10 MB. Compared with AdaBit, MBQuant achieves better performance and only has a negligible extra storage size cost.

\begin{table*}[ht]
\footnotesize
\caption{Results of ResNet-20 on CIFAR-100 dataset. ``Ind.'' indicates the individual training. ``Avg.'' indicates the average accuracy over all candidate bit-widths. ``$\dagger$'' indicates using the round-to-floor quantizer. ``$\ddagger$'' indicates using the center quantizer.}
\centering
\begin{tabular}{ccccccc}
\bottomrule[1.25pt]
\multirow{2}{*}{Networks} & \multirow{2}{*}{\begin{tabular}[c]{@{}c@{}}Bit-Widths \\ Size\end{tabular}} &  \multicolumn{5}{c}{Methods}  \\ \cline{3-7} 
                          &                             & Ind.  & APN  & AdaBit$\dagger$ & AdaBit$\ddagger$  & MBQuant (Ours)   \\ \hline\hline
\multirow{11}{*}{ResNet-20} & 4                           &    68.70    &  62.04 & 57.38 & 66.96 &  \textbf{69.41}  \\
                          & 3                           &   68.17     & 61.66  &  56.66 & 65.49 &  \textbf{69.28}   \\
                          & 2                           &   65.48     &  59.36    &   41.90   & 59.13    & 65.26     \\
                          & Avg.                           &  67.45 & 61.02 & 51.98 & 64.86 &   \textbf{67.98}   \\ 
                          & Size (MB)                           &   -    &  1.12 & 0.18    & 0.18  &  0.19     \\ \cline{2-7} 
                          & 8                           & 70.01      & 62.00   & 62.05  & 67.30 & \textbf{71.73}     \\
                          & 6                           & 69.62     & 61.81   & 62.39 & 67.31  & \textbf{71.35}    \\
                          & 4                           & 68.70     & 61.75   & 60.42 & 66.78 & \textbf{70.07}   \\
                          & 2                           &   65.48     &  58.77    & 18.72 & 57.54  &  \textbf{65.61}     \\
                          & Avg.                           & 68.45  & 61.08 & 50.90 &  64.73  & \textbf{69.69}  \\ 
                          & Size (MB)                          &    -  & 1.13     &  0.32  & 0.32  &  0.36     \\ \bottomrule[1.0pt]                          
\end{tabular}
\label{tab:cifar-100}
\end{table*}

\begin{table*}[!ht]
\footnotesize
\caption{Results of ResNet-20 on CIFAR-10 dataset. ``Ind.'' indicates the individual training. ``Avg.'' indicates the average accuracy over all candidate bit-widths. ``$\dagger$'' indicates using the round-to-floor quantizer. ``$\ddagger$'' indicates using the center quantizer.}
\centering
\begin{tabular}{ccccccc}
\bottomrule[1.25pt]
\multirow{2}{*}{Networks} & \multirow{2}{*}{\begin{tabular}[c]{@{}c@{}}Bit-Widths \\ Size\end{tabular}} &  \multicolumn{5}{c}{Methods}  \\ \cline{3-7} 
                          &                             & Ind.  & APN  & AdaBit$\dagger$  & AdaBit$\ddagger$  & MBQuant (Ours)   \\ \hline\hline
\multirow{11}{*}{ResNet-20}      & 4                           &    92.86    &  89.99    &  84.54 &  91.22 & \textbf{93.08}    \\
                          & 3                           &    92.12    &   89.55  &   84.25  & 91.11 & \textbf{92.60}    \\
                          & 2                           &   91.04     &   88.48   & 71.19 & 89.49 & \textbf{91.60}     \\ 
                          & Avg.                           & 92.01  & 89.34 &  79.99 & 90.61  &  \textbf{92.43}   \\
                          & Size (MB)                           &  -  &  1.10     &   0.16   & 0.16  & 0.17    \\\cline{2-7}  
                          & 8                           & 92.93      & 89.74    & 86.61 & 91.35 & \textbf{94.00}     \\
                          & 6                           & 92.91    & 89.65   & 86.81 & 91.22 & \textbf{93.61}    \\
                          & 4                           & 92.86     & 89.75   & 85.98 & 91.24 & \textbf{93.22}    \\
                          & 2                           &  91.04      & 88.67  & 56.84 & 88.99  &   \textbf{91.24}    \\ 
                          & Avg.                           & 92.44 & 89.45 & 79.06 &  90.70 &  \textbf{93.02}   \\ 
                          & Size (MB)                           &  -   &  1.11   &   0.30   & 0.30 &  0.34     \\ \bottomrule[1.0pt]                          
\end{tabular}
\label{tab:cifar-10}
\end{table*}

\subsection{Results on CIFAR-100/10}
As shown in Table\,\ref{tab:cifar-100}, our MBQuant outperforms the prior methods in terms of accuracy for different bit-width settings, with reduced storage overhead.
For instance, on ResNet-20, MBQuant respectively achieves accuracy gains of 6.13\%, 3.79\%, and 2.45\% for 2-, 3-, and 4-bit with the storage size of 0.19 MB, and 8.07\%, 3.29\%, 4.04\%, and 4.43\% for 2-, 4-, 6-, and 8-bit with the storage size of 0.36 MB. As a result, MBQuant obtains 3.12\% and 4.96\% gains on average accuracy for these two settings, respectively.
Notably, MBQuant even outperforms individual training in most bit-width cases and gives higher average accuracy. For example, compared with individual training, MBQuant presents 1.24\% average accuracy gains for 2-, 4-, 6-, and 8-bit settings. 
Moreover, AdaBit fails to converge for the 2-bit case if using the round-to-floor quantizer since such a quantizer destroys the zero-mean property of network weights, which is crucial for convergence~\cite{schoenholzdeep}. In contrast, using the center quantizer retains the performance well since it retains the zero-mean property of weights.


The results on CIFAR-10 are provided in Table\,\ref{tab:cifar-10}.
As can be seen, MBQuant achieves the best performance compared with APN~\cite{yu2021any} and AdaBit~\cite{jin2020adabits}.
Specifically, for ResNet-20 on CIFAR10, MBQuant respectively obtains 91.60\%, 92.60\%, and 93.08\% top-1 accuracy for 2-, 3-, and 4-bit, which corresponds improvements in accuracy by 2.11\%, 1.49\%, and 1.86\%, respectively. Similarly, MBQuant respectively obtains 91.24\%, 93.22\%, 93.61\%, and 94.00\% top-1 accuracy for 2-, 4-, 6-, and 8-bit, improving the accuracy by 2.25\%, 1.98\%, 2.39\%, and 2.65\%, respectively.
\begin{wrapfigure}{r}{0.55\textwidth} 
\centering
\includegraphics[width=\linewidth]{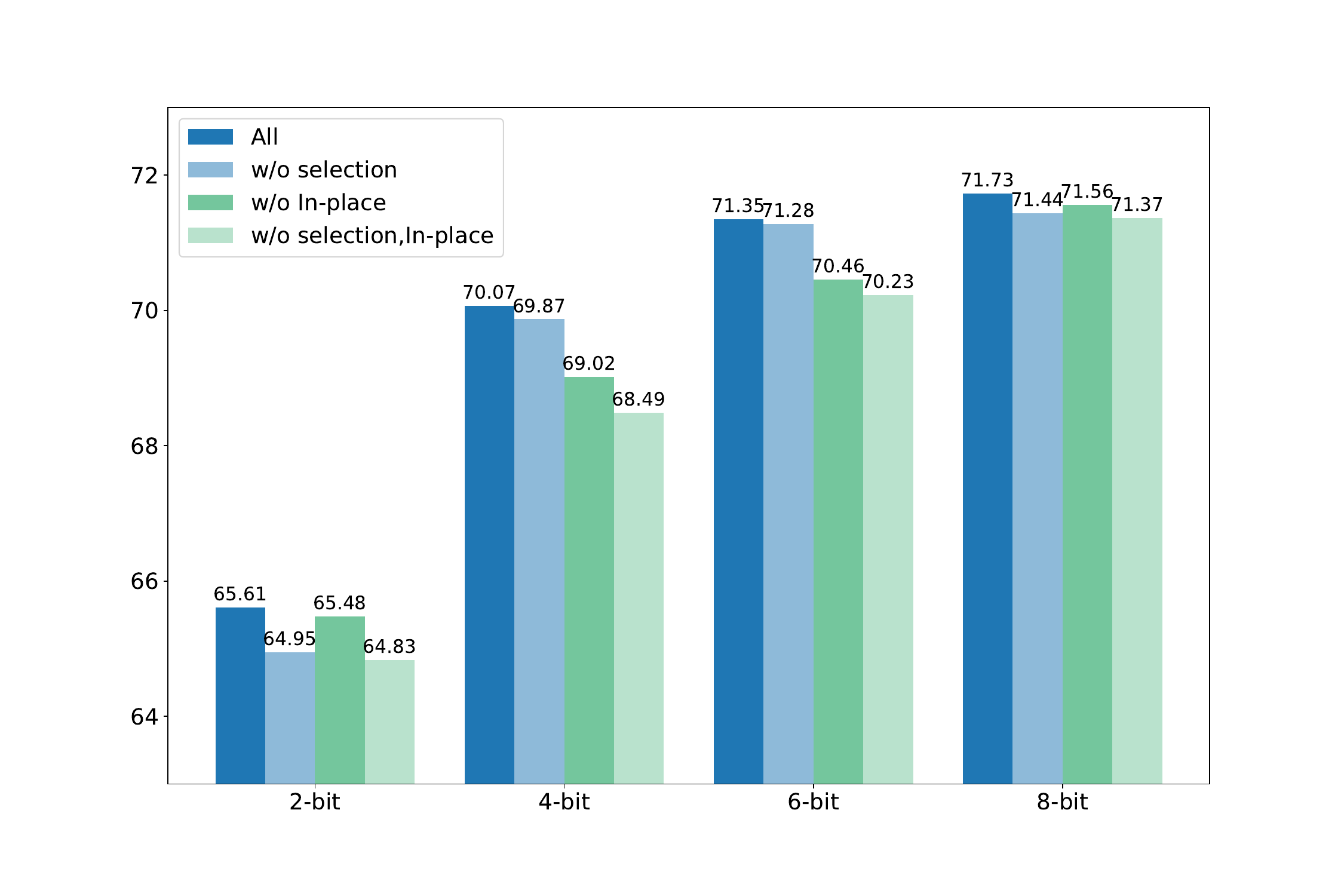} 
\caption{Influence of the amortization branch selection strategy and in-place distillation on the top-1 accuracy of ResNet-20 on CIFAR-100. ``w/o'' indicates without, ``selection'' indicates ``amortization branch selection strategy'', and ``In-place'' denotes ``in-place distillation''.}
\label{fig:ablation}
\end{wrapfigure}
Correspondingly, MBQuant improves the average accuracy by 1.82\% and 2.32\% for the aforementioned settings, respectively.
Thanks to the multi-branch topology, MBQuant reduces the storage requirement by storing weight in 2-bit. For ResNet-20 with 2-, 3-, and 4-bit settings, the storage cost is 0.17 MB. Compared with 0.16 MB of AdaBit, the extra storage overhead is 0.01 MB, which is negligible.
It is also worth noting that AdaBit which uses the round-to-floor quantizer suffers from performance degradation due to the round-to-floor function destroying the zero-mean property of network weight~\cite{schoenholzdeep}. It achieves only 71.19\% accuracy on the 2-bit case at 2-, 3-, and 4-bit settings. When using the center quantizer, AdaBit provides much better results. In contrast, our MBQuant does not require a change in the round function and successfully retains the accuracy of the 2-bit case.

\subsection{Ablation Study}

To validate the effectiveness of the proposed amortization branch selection strategy and in-place distillation, 
we conduct the ablation study with ResNet-20 on CIFAR-100, at the 2-, 4-, 6-, and 8-bit settings. As illustrated in Fig.\,\ref{fig:ablation}, the accuracy loss occurs either by removing the amortization branch selection strategy or in-place distillation, which demonstrates their effectiveness. 
Specifically, for the smallest bit-width, removing the amortization branch selection strategy causes a considerable degradation, \emph{i.e.}, 0.66\%. Such results demonstrate the importance of our amortization branch selection strategy for the smallest bit-width. 
Removing in-place distillation respectively leads to the 0.13\%, 1.05\%, 0.89\%, and 0.17\% accuracy degradation for 2-, 4-, 6-, and 8-bit cases, proving its effectiveness. Moreover, if the amortization branch selection strategy and in-place distillation are both removed, the largest accuracy degradation occurs for all bit-widths, in which the accuracy drops 0.78\%, 1.58\%, 1.12\%, and 0.36\% for 2-, 4-, 6-, and 8-bit, respectively.



\section{Conclusion}

This paper addresses the issue of arbitrary bit-width QNNs by introducing a novel method called MBQuant. We begin by demonstrating that existing methods for arbitrary bit-width QNNs suffer from significant inherent quantization error from switching weight bit-widths and switching activations bit-widths, hindering their performance. To overcome this limitation, MBQuant utilizes a multi-branch topology. Specifically, MBQuant duplicates the network into multiple independent branches, each of which has fixed 2-bit weights and the activations remain the input bit-width. The computation of a desired bit-width is enabled by combining the output of these branches, thereby eliminating quantization errors that arise from switching weight bit-widths.
In addition, an amortization branch selection strategy is introduced to amortize quantization error from switching activations bit-widths into branches for improving the performance of the smallest branch. Finally, an in-place distillation strategy is adopted to further enhance MBQuant. 
Extensive experiments conducted over various datasets, bit-widths, and networks demonstrate the effectiveness of MBQuant.






\bibliographystyle{elsarticle-num-names} 
\bibliography{main}





\end{document}